\documentclass[conference]{IEEEtran}
\IEEEoverridecommandlockouts
\usepackage{cite}
\usepackage{amsmath,amssymb,amsfonts}
\usepackage{algorithmic}
\usepackage{graphicx}
\usepackage{textcomp}
\usepackage{xcolor}
\usepackage{graphicx}
\usepackage{float}
\usepackage{caption}
\usepackage{placeins}
\def\BibTeX{{\rm B\kern-.05em{\sc i\kern-.025em b}\kern-.08em
    T\kern-.1667em\lower.7ex\hbox{E}\kern-.125emX}}
\begin{document}

\title{When Do LLMs Actually Help With Data Quality? Evaluating LLMs as Data Quality Annotators}

\author{
\IEEEauthorblockN{Praphulla Lal Shrestha}
\IEEEauthorblockA{
\textit{Independent Researcher} \\
Kathmandu, Nepal \\
praphullashrestha@gmail.com
}
}

\maketitle

\begin{abstract}
\normalfont
LLMs have been increasingly used to catch data quality issues automatically, but we know very little about how consistent these judgments actually are. This study tests an LLM on two e-commerce data quality tasks, entity matching and brand mislabeling, against rule based baselines and human verified ground truth, under both zero-shot and few-shot prompting. On entity matching while using the Abt Buy benchmark (2,194 labeled pairs), a simple rule based baseline (F1=0.950) performed about as well as LLM zero shot prompting (F1=0.948). Moreover, a few-shot prompt revision that looked effective on a small validation sample reduced full-scale performance to F1=0.914. This showed that small sample prompt evaluation can be misleading. On brand mislabeling detection, using 500 Amazon product listings with synthetically injected labeling errors, the LLM clearly outperformed a naive rule based baseline (F1=0.833 vs 0.721), because it could draw on background knowledge of brand product relationships that a simple rule could not access. Testing consistency across repeated runs (200 pairs, 5 runs at temperature 0.7) showed the model agreeing with itself 99.7\% of the time on average, with 99\% of pairs giving identical answers across all 5 runs. Using majority voting across these runs only improved F1 by 0.005, at 5 times the inference cost. These results suggest that the value of using an LLM over traditional methods depends heavily on the task. LLMs offer little advantage when strong lexical signals already exist, but a clear advantage when the task requires background knowledge, all while remaining highly consistent across repeated queries.
\end{abstract}

\section{Introduction}

The quality of large-scale structured data has become increasingly important as organizations rely heavily on web-scraped and e-commerce datasets for search, recommendation systems, analytics, and business decision-making. However, ensuring data quality is maintained remains a big challenge. Common data issues such as duplicate records, incorrect brand assignments, inconsistent metadata, missing values frequently arise during data collection and integration. Although manual review and rule based validation techniques are widely used to identify these problems, they become very difficult to scale as datasets grow from few hundreds to millions of records. In industrial data pipelines, where millions of records require continuous monitoring, there is a growing need for automated approaches that can make reliable data quality judgements with least amount of manual intervention.

Large Language Models (LLMs) have recently emerged as a possible solution for such tasks because of their ability to understand natural language and generalize beyond explicit rules. Consequently, they are increasingly being explored as flexible annotators for applications such as entity matching, data labeling, record validation, and anomaly detection \cite{gilardi2023,wang2024,aguda2024}. Despite this growing adoption, most evaluations emphasize one-time predictive performance but give comparatively little attention to the reliability of the annotations themselves \cite{alvarado2025}. In practical data quality workflows, an annotator that produces different outputs for the same input across repeated executions can definitely reduce confidence in automated decision making. Therefore, measuring accuracy only is not enough, it is important to assess whether LLMs produce consistent judgments under repeated inference.

This study evaluates the reliability of LLMs as data quality annotators through two representative e-commerce data quality tasks: entity matching and brand mislabeling detection, using GPT-4o-mini as the model under test. Entity matching experiments are conducted on the publicly available Abt-Buy benchmark, while brand mislabeling is evaluated using a synthetically corrupted Amazon product dataset. For each task, LLM-based approaches are compared against established rule-based baselines. Along with comparing zero-shot and few-shot prompting strategies, repeated inference experiments are performed to quantify the consistency of LLM predictions across multiple independent runs.

The results show that the value of LLMs depends strongly on the characteristics of the data quality task. For entity matching, zero shot LLM prompting achieves performance comparable to a simple Jaccard similarity baseline which suggests limited advantage over traditional methods for this problem. In contrast, the LLM substantially outperforms a rule-based approach in detecting brand mislabeling which demonstrates the benefit of contextual and semantic reasoning where explicit rules are insufficient. Consistency experiments further reveal that the model produces highly stable predictions across repeated runs. Overall, these findings provide practical evidence on when LLMs are a good choice for automated data quality annotation and highlight important considerations when evaluating prompt-based systems.

\FloatBarrier
\newpage
\begin{figure}[t]
\centering
\includegraphics[width=\columnwidth]{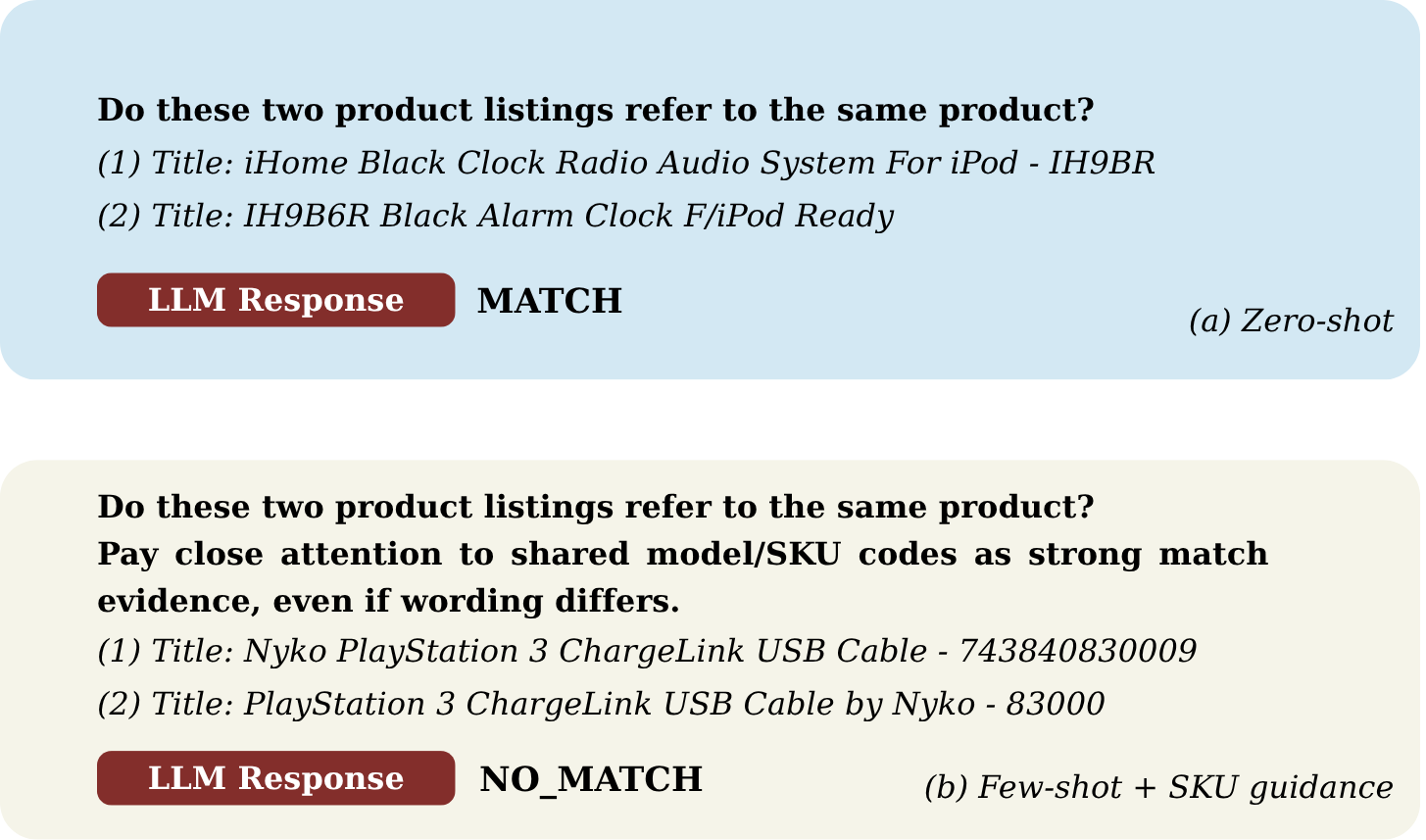}
\caption{Example prompts and responses for the two prompting strategies tested. Panel (a) shows a correctly resolved match under zero-shot prompting; panel (b) shows a true match incorrectly rejected under few-shot SKU-guidance prompting.}
\label{fig:prompt_examples}
\end{figure}

\section{Related Work}

\textbf{Classical entity resolution.} Entity resolution, linking records that refer to the same real world entity without a shared identifier, predates LLM based approaches by decades. Binette and Steorts \cite{binette2022} trace the field from statistical record linkage through deterministic, probabilistic, and clustering-based methods. They document deterministic matching as a long standing, interpretable baseline that degrades as attribute noise increases. This motivated richer probabilistic approaches such as the Fellegi-Sunter model. This grounds the Jaccard baseline used in the present study as an established method and explains the low overlap failure cases found in Section 4.1.

\textbf{LLM-based entity matching.} Wang et al. \cite{wang2024} compare pairwise ``matching,'' pairwise ``comparing,'' and listwise ``selecting'' prompting strategies, finding that exploiting relationships across candidate records substantially improves F1 (comparing beats matching by 10.7 points; selecting improves on comparing by a further 5.3). Their proposed COMEM framework combines models of different strengths to improve F1 by 2 to 18\% at lower cost. The present study uses only the plain pairwise ``matching'' strategy, so \cite{wang2024}'s critique that independent classification ignores global consistency (e.g., one-to-one mapping) applies directly here and is revisited in Section 5. More directly comparable, since it evaluates the same Abt-Buy benchmark, Bopardikar et al. \cite{bopardikar2025} test multi-step reasoning prompts and a debate-based alternative. Their zero-shot baseline F1 (0.849) is notably lower than the 0.948 obtained here, likely because \cite{bopardikar2025} uses the full, heavily imbalanced DeepMatcher split (1,028 positive vs. 8,547 negative pairs) rather than the balanced 2,194-pair set built for this study. Bopardikar et al. \cite{bopardikar2025} also found added reasoning complexity helps zero-shot but hurts few-shot prompting on this dataset, closely mirroring the present study's own few-shot generalization failure (Section 4.1), suggesting added prompting complexity does not reliably help once baseline signal is already present.

\textbf{Product attribute and brand data quality.} Wang et al. \cite{wang2020} address validating catalog attribute values (e.g., brand), motivated by the same concern as the present study's second task, and explicitly note that naive keyword presence checking fails when a value is textually present but assigned to the wrong attribute, precisely the failure mode behind the low precision (0.564) of the rule-based baseline in Section 4.2. Their proposed MetaBridge model requires training and per category unlabeled data, unlike the training free zero-shot prompting used here, which reaches comparable correctness detection (F1=0.833) with no training at all. A more recent auto-prompt cascade approach \cite{autoprompt2025} bootstraps category specific instructions from a small seed set, improving Claude 3.5 Sonnet's correctness F1 from 67.37\% to 91.13\% while cutting human instruction writing effort by roughly 99\%. The present study's generic zero-shot prompt sits closer to \cite{autoprompt2025}'s baseline tier than its refined cascade, suggesting category specific auto-generated prompting is a concrete direction for improving brand mislabeling detection further (Section 6).

\textbf{LLMs as data annotators.} Aguda et al. \cite{aguda2024} evaluate GPT-4, PaLM 2, and MPT Instruct as financial relation extraction annotators against expert and individual crowdworker labels, finding LLMs substantially outperform non expert crowdworkers (GPT-4: 68.4\% vs. 38.6\% for MTurk). Notably, no model in \cite{aguda2024}, including GPT-4, achieved perfect self-agreement across two identical runs, and prompt choice affected consistency more than temperature did. This contrasts with the present study's near-perfect self-agreement (99.7\%) on a simpler binary task, supporting the view developed in Section 5 that consistency depends on task difficulty and label ambiguity, not model capability alone.

\textbf{Reliability and consistency under repeated sampling.} Sharma et al. \cite{sharma2026} study cognitive distortion detection, where human ground truth is itself unreliable (33.7\% expert agreement), and treat cross run consistency as a reliability proxy, finding the majority label stable across 5 runs for over 84\% of examples. This differs from the present study, whose ground truth is objectively verified rather than subjective, so consistency here is a confirmatory signal rather than a substitute for ground truth. Alvarado Gonzalez et al. \cite{alvarado2025} show single run LLM evaluations produce unstable rankings (83\% of conditions showed a ranking inversion against a 3 run majority vote) and recommend at least two repeated runs as a practical minimum. A separate causal study of temperature in LLM as a judge settings \cite{temperature2025} finds consistency degrades sharply at high temperature, driven substantially by output format errors rather than declining judgment quality, an effect the present study's simple binary output format may have limited, since zero unparseable responses occurred even at temperature 0.7. Both studies motivate the limitation, discussed in Section 5, that consistency here was measured at only one non zero temperature point rather than a full sweep.

\textbf{Cost-efficient LLM pipelines and small model evaluation.} Schnabel et al. \cite{schnabel2025} show a cheap, structurally decomposed relevance assessment pipeline (a binary filter followed by refinement) can outperform a single expensive flagship model call, improving even GPT-4o's own accuracy by 9.7\%. This contrasts with the present study’s own attempts at complexity, where a few shot prompt revision and 5-runs majority voting each added cost for marginal or negative return. Han et al. \cite{han2025} similarly find, across six LLMs and five prompting strategies for literature review screening, that self reflection prompting consistently underperforms simpler prompts, a third independent instance of added sophistication failing to help. Han et al. \cite{han2025} also find GPT-4o-mini competitive with larger models at a fraction of the cost, which supports the present study's model choice, corroborated by OpenAI's own benchmarks \cite{openai2024} showing GPT-4o-mini outperforming similarly priced models like Gemini Flash and Claude Haiku on general reasoning (82.0\% vs. 77.9\% MMLU).

\section{Methodology}

\subsection{Datasets}

For entity matching, this study uses the Abt-Buy benchmark, consisting of Abt.csv (1,081 products) and Buy.csv (1,092 products), along with a human verified perfect mapping file identifying true matches between the two catalogs. From this, a labeled pairs set of 2,194 rows was constructed: positive pairs drawn directly from the mapping file, and negative pairs generated through random non matching sampling, producing a class balanced dataset.

For brand mislabeling, this study uses Amazon.csv (from the Amazon-Google Products benchmark), sampled down to 500 rows with a populated manufacturer field. Since no existing ground truth for mislabeling exists in this dataset, ground truth was constructed synthetically: approximately half of the rows had their manufacturer field swapped with a different, real manufacturer drawn from elsewhere in the dataset, while the remaining rows were left untouched, producing a labeled correctness dataset with a known verifiable ground truth.

\subsection{Baselines}

For entity matching, the baseline is a Jaccard word overlap similarity measure computed on normalized product names (lowercased, punctuation stripped), where matches are determined by a similarity threshold. A threshold of 0.2 was found to give the best F1 score among the thresholds tested and is used as the reported baseline throughout this study.

For brand mislabeling, the baseline is a naive substring check: a product is flagged as correctly labeled if the manufacturer name appears within the product title, and flagged as mislabeled otherwise.

\subsection{LLM Setup}

All LLM experiments use GPT-4o-mini via the OpenAI API. Zero-shot prompting is used as the primary method for both tasks, and temperature is set to 0 for all main results to ensure reproducibility. For entity matching, a few-shot prompting variant is additionally tested in which the prompt includes explicit guidance to prioritize shared model or SKU codes as strong evidence of a match, along with two worked examples. Consistency testing is conducted only for the entity matching task, using five repeated runs at temperature 0.7 on a 200-pair stratified subsample.

\subsection{Evaluation}

All methods are evaluated using precision, recall, F1, and accuracy against ground truth. For consistency testing, two additional metrics are used: the per-pair agreement rate across the five repeated runs and the accuracy of a majority-vote prediction taken across those five runs.

\section{Results}

\subsection{Entity Matching (Abt-Buy, 2,194 Pairs)}

\begin{table}[htbp]
\caption{Precision, recall, F1, and accuracy for the three entity matching methods evaluated on the Abt-Buy benchmark (2,194 labeled pairs).}
\label{tab:entity_matching}
\centering
\begin{tabular}{lcccc}
\hline
Method & Precision & Recall & F1 & Accuracy \\
\hline
Rule-based (Jaccard, threshold=0.2) & 0.994 & 0.910 & 0.950 & 0.952 \\
LLM zero-shot & 0.999 & 0.902 & 0.948 & 0.951 \\
LLM few-shot + SKU guidance & 0.994 & 0.846 & 0.914 & 0.920 \\
\hline
\end{tabular}
\end{table}

\begin{figure}[!htbp] \centering \includegraphics[width=\columnwidth]{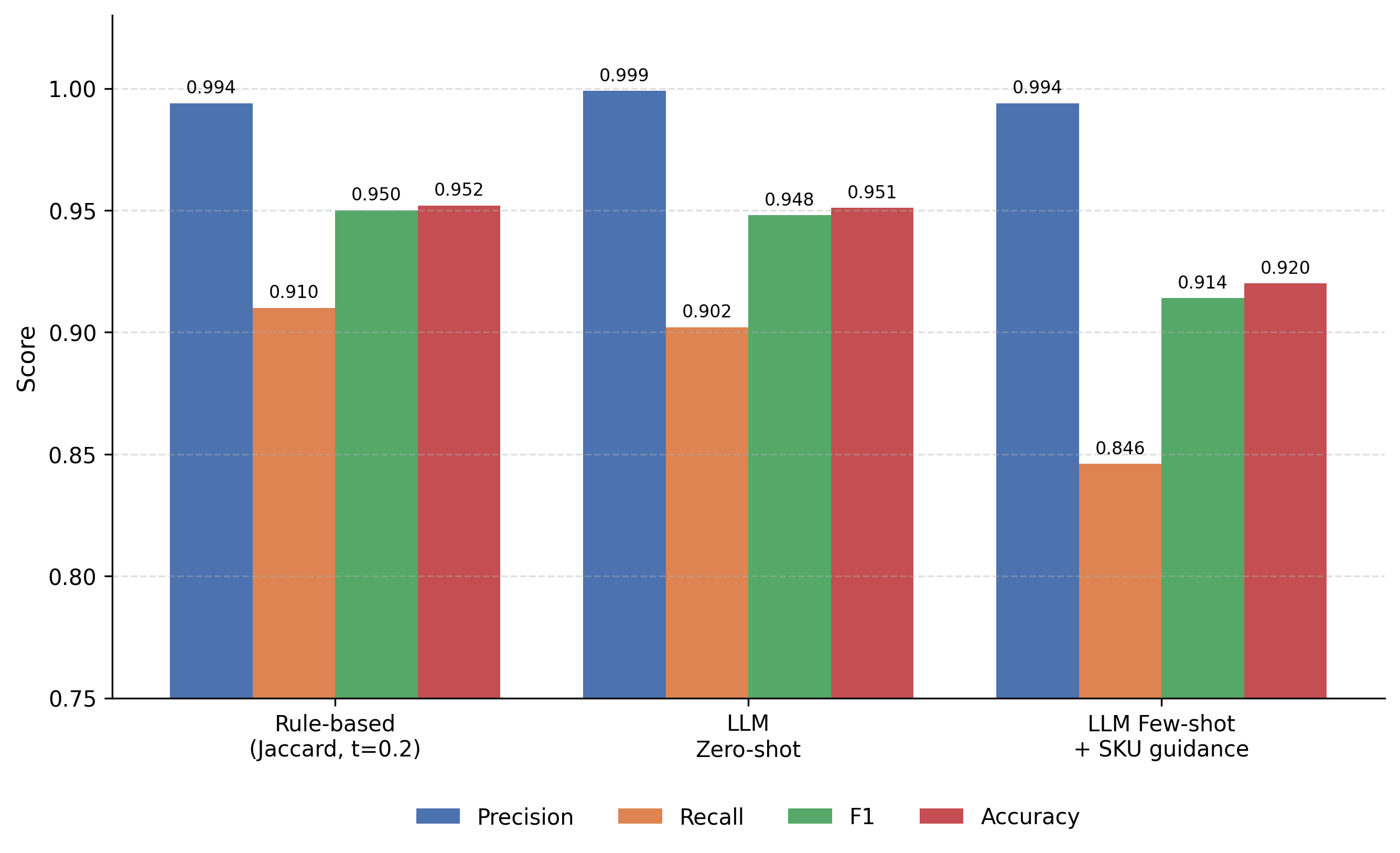} \caption{Comparison of precision, recall, F1, and accuracy across the three entity matching methods on the Abt-Buy benchmark.} \label{fig:entity_matching_results} \end{figure}

An error analysis was conducted comparing the zero-shot LLM predictions against the rule-based baseline. Of the 2,194 pairs, both methods agreed and were correct on 2,022 pairs. The rule-based method was correct while the LLM was wrong on 67 pairs, the LLM was correct while the rule-based method was wrong on 64 pairs, and both methods were wrong on 41 pairs. LLM wins mainly occurred on pairs with low lexical overlap but clear semantic equivalence, such as abbreviated retail strings or reformatted SKU codes. Rule-based wins occurred mainly on pairs sharing an exact alphanumeric model or SKU code but differing descriptive text. These were the cases where the LLM underweighted a strong exact match signal.

A prompt revision explicitly instructing the model to prioritize SKU or code matches was validated on a small sample consisting of the 67 previously wrong cases plus 50 control cases, fixing 28 of the 67 errors while breaking only 1 of the 50 control cases. At full scale across all 2,194 pairs, however, this same prompt reduced F1 from 0.948 to 0.914, with recall dropping from 0.902 to 0.846. The intervention appears to have over generalized, causing the model to become more code dependent across the broader dataset than the small validation sample suggested.

\subsection{Brand Mislabeling (Amazon, 500 Listings)}

\begin{table}[htbp]
\caption{Precision, recall, F1, and accuracy for the two brand mislabeling detection methods evaluated on 500 Amazon product listings.}
\label{tab:brand_mislabeling}
\centering
\begin{tabular}{lcccc}
\hline
Method & Precision & Recall & F1 & Accuracy \\
\hline
Rule-based (manufacturer-title) & 0.564 & 1.000 & 0.721 & 0.622 \\
LLM zero-shot & 0.827 & 0.840 & 0.833 & 0.836 \\
\hline
\end{tabular}
\end{table}

The same error analysis was conducted for the brand mislabeling task. Of the 500 listings, both methods agreed and were correct on 272 listings. The LLM was correct while the rule-based method was wrong on 146 listings, both methods were wrong on 43 listings, and the rule-based method was correct while the LLM was wrong on 39 listings. The rule-based method's high recall but low precision stems from flagging any product whose manufacturer name does not literally appear in the title, even when the label is correct. For example, ``ZoneAlarm Anti-Spyware'' is manufactured by ``Zone Labs,'' and ``SYMC Backup Exec'' is manufactured by ``Symantec,'' where SYMC is a stock ticker abbreviation rather than the full company name. The LLM correctly resolves these cases by drawing on background knowledge of brand relationships. Conversely, the LLM's errors mostly involved accepting unfamiliar or niche publisher names as plausible without verification, missing genuine mislabels that the stricter rule-based check caught.

\begin{center}
\includegraphics[width=\columnwidth]{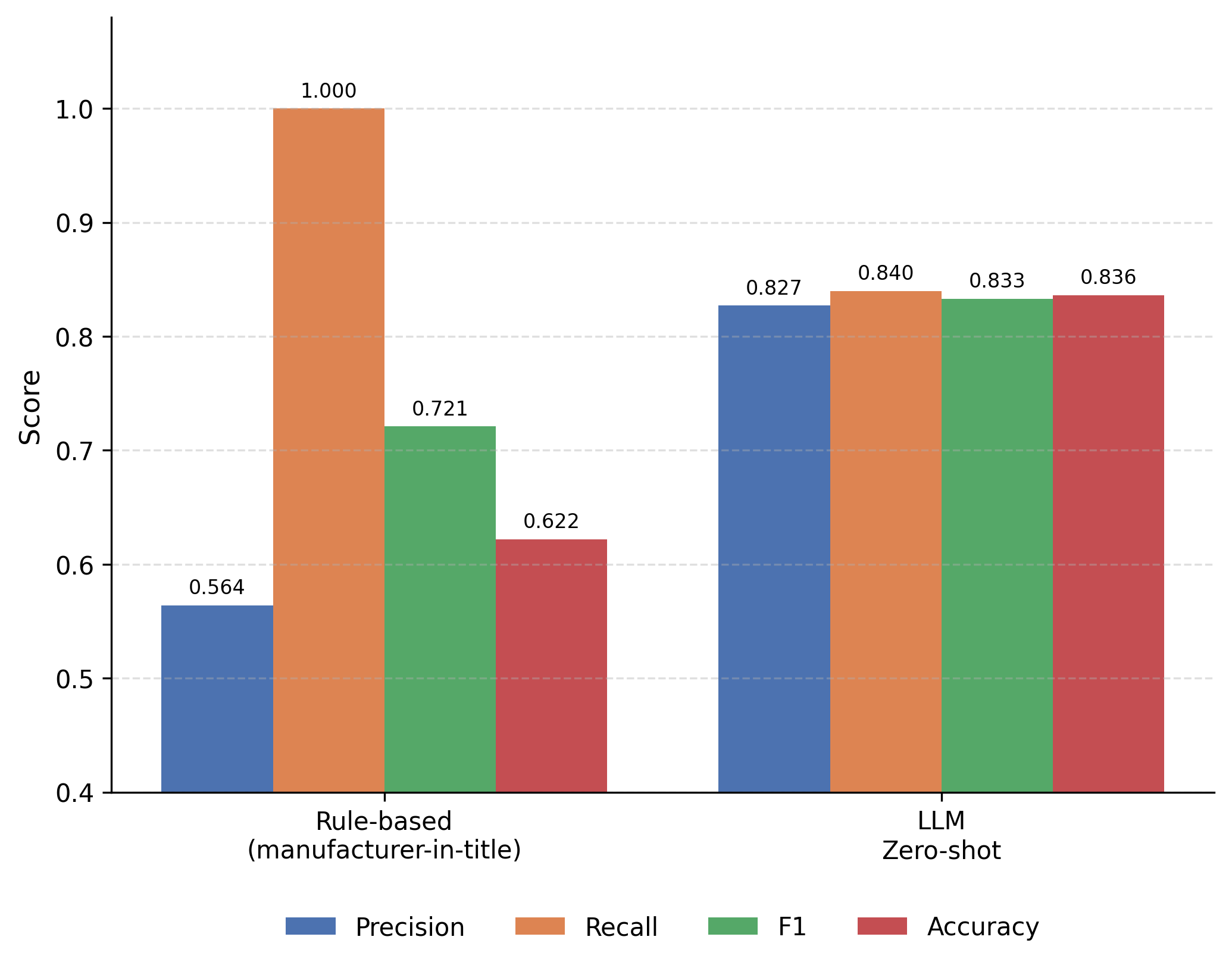}
\captionof{figure}{Precision, recall, F1, and accuracy for the two brand mislabeling detection methods evaluated on 500 Amazon product listings.}
\label{fig:brand_mislabeling_results}
\end{center}

\subsection{Consistency and Reliability (200-Pair Sample, 5 Runs, Temperature = 0.7)}
Across five repeated runs at temperature 0.7 on a 200-pair stratified sample, the average per-pair agreement rate was 0.997, and 99.0\% of pairs produced identical answers across all five runs. A majority vote taken across the five runs achieved an F1 of 0.974, compared to an F1 of 0.969 for a single run at temperature 0 on the same 200 pair sample. The LLM showed high self-consistency even under non-zero temperature. Majority vote ensembling across five runs yielded only a small F1 improvement of 0.005 at five times the inference cost, suggesting limited practical benefit from ensembling for this task given the already high single run consistency.

\begin{figure}[!htbp]
\centering
\includegraphics[width=\columnwidth]{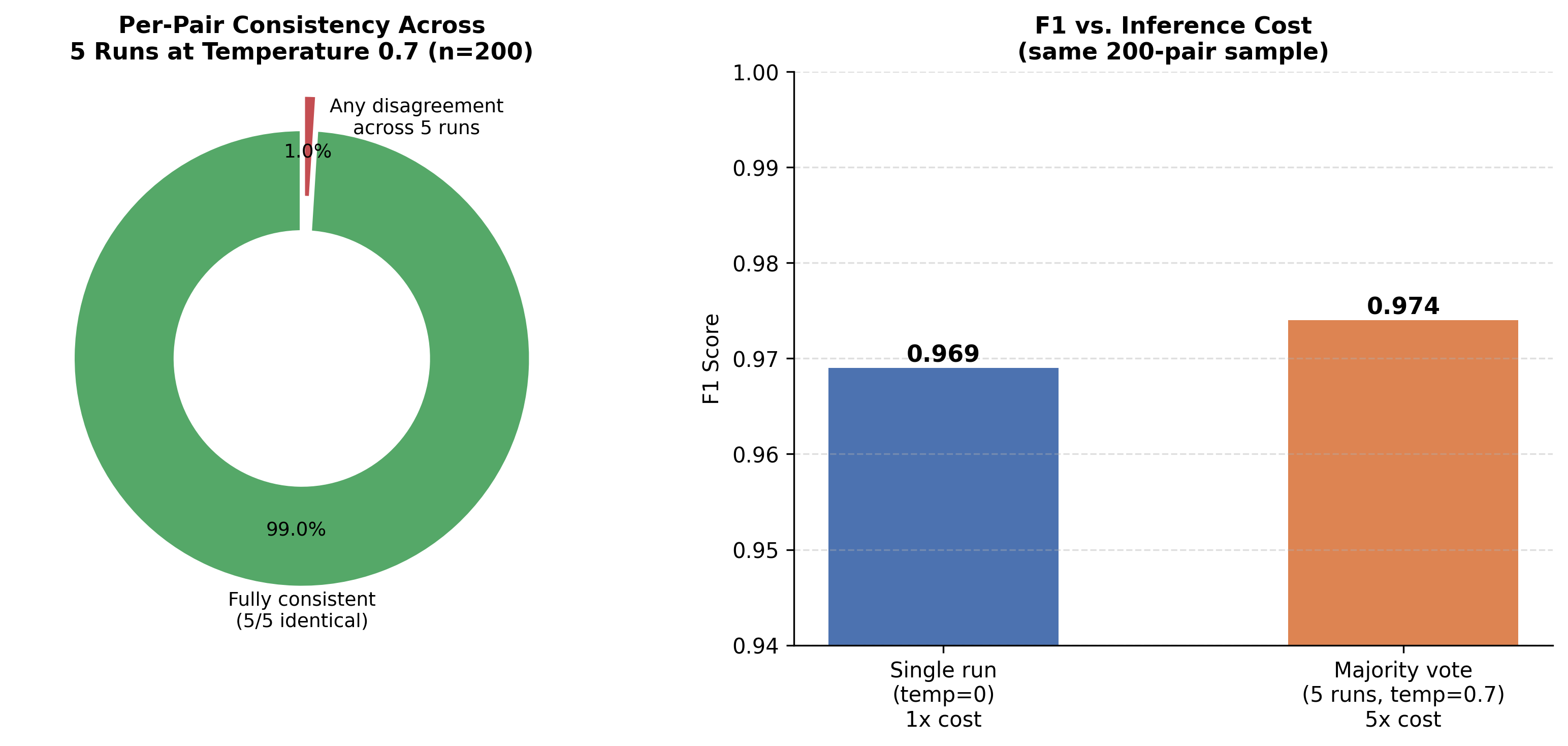}
\caption{Consistency of LLM predictions across five repeated runs at temperature 0.7 on a 200-pair sample (left), and F1 score for a single deterministic run versus a five-run majority vote on the same sample (right).}
\label{fig:consistency_results}
\end{figure}

\section{Discussion}

The central finding of this study is that the value of LLM based annotation over rule based methods is strongly task dependent. On entity matching, a simple Jaccard baseline (F1=0.950) essentially matched LLM zero shot prompting (F1=0.948), since Abt-Buy product names are largely literal with high lexical overlap between matches, leaving little room for semantic understanding to add value. Brand mislabeling told a different story. The rule-based baseline reached only F1=0.721 (precision 0.564, recall 1.000), while the LLM reached 0.833 (precision 0.827, recall 0.840), because many correct manufacturer labels are not literally present in the product title (e.g., an abbreviation or subsidiary brand name), a case a string-matching rule cannot resolve but an LLM can, drawing on background knowledge of brand relationships \cite{wang2020,autoprompt2025}. LLMs, then, offer the greatest advantage on tasks requiring external or contextual knowledge not recoverable from the text alone, and little advantage when a simple lexical heuristic already captures most of the signal.

A second, methodologically important finding came from an attempted prompt improvement for entity matching. A revised prompt that instructed the model to prioritize exact SKU/model codes was validated on a small set (67 previously wrong cases plus 50 control cases). This approach fixed 28 of the 67 errors while breaking only 1 of the 50 controls, which is a clear win. At full scale across all 2,194 pairs however, this same prompt reduced F1 from 0.948 to 0.914 (recall fell from 0.902 to 0.846). The model became more code-dependent across the broader dataset than the small sample suggested. This is a useful caution for LLM based system development. Validating a prompt change on a small, hand picked sample can substantially overstate its benefit. Bopardikar et al. \cite{bopardikar2025} found the same directional effect (added reasoning complexity helped zero-shot but hurt few-shot prompting on this same dataset), and Han et al. \cite{han2025} separately found self-reflection prompting underperforms simpler prompts, together suggesting added complexity helps only when well targeted and validated at scale.

Reliability, in the sense of run to run consistency, was high across five runs at temperature 0.7 on a 200 pair sample: the model agreed with itself 99.7\% of the time on average and gave identical answers across all five runs for 99\% of pairs. Majority voting across the five runs gave only a marginal gain over a single run (F1 0.974 vs. 0.969) at five times the cost, so a single zero-shot run at temperature 0 is a reasonable practical choice here. This should be read narrowly, however as other work has found much lower self agreement on broader or more subjective tasks. For example, no perfect self agreement was observed even across identical runs for financial relation extraction \cite{aguda2024}, and GPT-4o-mini falling toward the less consistent end of tested models on math reasoning \cite{alvarado2025}. The narrow, binary nature of the present task likely makes it inherently easier to answer consistently than these more open-ended tasks, supporting the view that consistency depends on task difficulty and label ambiguity, not model capability alone. Consistency here was also measured at only one non-zero temperature point; a full temperature sweep elsewhere \cite{temperature2025} shows consistency degrading progressively as temperature rises, driven substantially by output formatting errors, a factor likely muted here given the present task's simple binary output format produced zero unparseable responses even at temperature 0.7.

\section{Limitations}

The Abt-Buy benchmark is a comparatively easy entity matching dataset with structurally consistent names across catalogs. Due to this reason results may not generalize to noisier real world catalogs. The brand mislabeling ground truth was constructed synthetically via random manufacturer swaps, which tends to produce clearly implausible mismatches rather than the subtler errors seen in practice. Only a single LLM (GPT-4o-mini) was evaluated, so the reliability and relative advantage findings reported here may not generalize to other model families or sizes. Consistency testing was conducted only for the entity matching task, leaving the brand mislabeling task's run to run reliability untested. Both tasks were evaluated as independent, per pair or per row classification problems, without considering consistency across related records, a limitation shared with much of the existing entity matching literature \cite{wang2024}. Finally, the zero shot prompts used in both tasks were generic rather than tailored to specific product categories or attribute types; recent work on automatically generating category specific prompts for catalog quality tasks \cite{autoprompt2025} shows substantially higher performance than generic prompting, suggesting the brand mislabeling results reported here likely represent a lower bound on what LLM based approaches could achieve with more tailored prompting.

\section{Conclusion}

This study examined how reliable LLM based annotation is for data quality tasks, evaluating GPT-4o-mini against rule-based baselines on two e-commerce data quality problems, entity matching and brand mislabeling detection, and separately testing how consistent its predictions are across repeated runs. The results show that the value of using an LLM over a traditional rule-based method depends on the nature of the task: zero-shot LLM prompting performed on par with a simple word overlap baseline for entity matching (F1 0.948 vs. 0.950), but substantially outperformed a rule-based baseline for brand mislabeling detection (F1 0.833 vs. 0.721), where background knowledge of brand relationships gave the LLM a clear advantage a lexical rule could not replicate. The model was also highly self-consistent across repeated runs, agreeing with itself on 99\% of cases across five independent trials, though a targeted prompt improvement that appeared successful on a small validation sample failed to generalize when applied to the full dataset, underscoring the need to validate prompt changes at full scale rather than on small, hand-picked samples before treating them as reliable improvements.

Taken together, these findings suggest that practitioners should evaluate LLM-based data quality annotation on a per task basis rather than assuming a uniform benefit over existing methods, and should be cautious about generalizing conclusions from small scale prompt testing. Future work should extend this evaluation to additional model families, test consistency on the brand mislabeling task, and explore harder, more realistic mislabeling scenarios closer to the kinds of subtle brand and category conflicts encountered in production scale e-commerce data pipelines.


\begin{thebibliography}{00}

\bibitem{gilardi2023}
F. Gilardi, M. Alizadeh, and M. Kubli, ``ChatGPT outperforms crowd-workers for text-annotation tasks,'' \textit{Proceedings of the National Academy of Sciences}, vol. 120, no. 30, 2023.

\bibitem{wang2024}
C. Wang et al., ``Match, Compare, or Select? An Investigation of Large Language Models for Entity Matching,'' arXiv:2405.16884, 2024.

\bibitem{aguda2024}
T. Aguda, S. Siddagangappa, E. Kochkina, et al., ``Large Language Models as Financial Data Annotators: A Study on Effectiveness and Efficiency,'' arXiv:2403.18152, 2024.

\bibitem{alvarado2025}
A. Alvarado Gonzalez et al., ``Do Repetitions Matter? Strengthening Reliability in LLM Evaluations,'' arXiv:2509.24086, 2025.

\bibitem{binette2022}
O. Binette and R. C. Steorts, ``(Almost) All of Entity Resolution,'' \textit{Science Advances}, vol. 8, no. 1, 2022.

\bibitem{bopardikar2025}
N. Bopardikar, H. Wang, and J. Zou, ``Structured Multi-Step Reasoning for Entity Matching Using Large Language Model,'' Arizona State University, preprint.

\bibitem{wang2020}
C. Wang, X. Xu, B. Li, et al., ``Automatic Validation of Textual Attribute Values in E-commerce Catalog by Learning with Limited Labeled Data,'' in \textit{Proc. KDD}, 2020. arXiv:2006.08779.

\bibitem{autoprompt2025}
``Auto Prompting Without Training Labels: An LLM Cascade for Product Quality Assessment in E-commerce Catalogs,'' arXiv:2510.23941, 2025.

\bibitem{sharma2026}
D. Sharma, A. Agarwal, and K. Sirts, ``Towards Consistent Detection of Cognitive Distortions: LLM-Based Annotation and Dataset-Agnostic Evaluation,'' in \textit{Proc. LREC}, 2026. arXiv:2511.01482.

\bibitem{temperature2025}
``The Necessity of Setting Temperature in LLM-as-a-Judge,'' arXiv:2603.28304, 2025.

\bibitem{schnabel2025}
J. A. Schnabel, J. R. Trippas, F. Scholer, and D. Hettiachchi, ``Multi-stage Large Language Model Pipelines Can Outperform GPT-4o in Relevance Assessment,'' in \textit{Proc. WebConf}, 2025.

\bibitem{han2025}
Z. Han, A. Mathrani, and T. Susnjak, ``Evaluating Prompting Strategies and Large Language Models in Systematic Literature Review Screening: Relevance and Task-Stage Classification,'' Massey University, 2025.

\bibitem{openai2024}
OpenAI, ``GPT-4o mini: Advancing Cost-Efficient Intelligence,'' 2024.

\end{thebibliography}
\end{document}